\documentclass[journal=acscii,manuscript=article,layout=singlecolumn,email=True]{achemso}

\usepackage[utf8]{inputenc}
\usepackage{amsmath}
\usepackage{mathtools}
\usepackage{amssymb}
\usepackage{color}
\usepackage{units}
\usepackage{bbm}

\usepackage{tikz}
\usepackage{graphicx}% Include figure files
\usepackage{dcolumn}% Align table columns on decimal point
\usepackage{bm}% bold math
\usepackage{upgreek}
\usepackage{array}
\usepackage{booktabs}
\usepackage{microtype}
\usepackage{cmap}
\usepackage{siunitx}
\usepackage{hyperref}
\usepackage{notoccite}
\usepackage[nodayofweek]{datetime}
\usepackage{expl3}
\usepackage{tabulary}
\usepackage{outlines}

\usepackage{caption}
\newcommand{\rxn}{\mathrm{rxn}} 
\newcommand{\sub}{\mathrm{sub}} 
\newcommand{\tot}{\mathrm{tot}} 
 
\newcommand{\roughly}{\raise.17ex\hbox{$\scriptstyle\mathtt{\sim}$}}

\newcommand*\pct{\protect\scalebox{0.9}{\%}}

\DeclareMathOperator*{\argmin}{arg\,min}
 
\author{John S. Schreck}
\email{jsschreck@gmail.com}
\affiliation{Department of Chemical Engineering, Columbia University, New York, NY 10027, USA}

\author{Connor W. Coley}
\affiliation{Department of Chemical Engineering, Massachusetts Institute of Technology, Cambridge, MA 02139, USA}

\author{Kyle J. M. Bishop}
\email{kyle.bishop@columbia.edu}
\affiliation{Department of Chemical Engineering, Columbia University, New York, NY 10027, USA}

\title{Learning retrosynthetic planning through self-play}


\hypersetup{colorlinks, citecolor={blue}, linkcolor={red}, urlcolor={violet}, pdftitle={}, pdfauthor={J. S. Schreck, C. W. Coley, K. J. M. Bishop}, pdfdisplaydoctitle}

\begin{document}

\begin{abstract}

The problem of retrosynthetic planning can be framed as one player game, in which the chemist (or a computer program) works backwards from a molecular target to simpler starting materials though a series of choices regarding which reactions to perform. This game is challenging as the combinatorial space of possible choices is astronomical, and the value of each choice remains uncertain until the synthesis plan is completed and its cost evaluated.   Here, we address this problem using deep reinforcement learning to identify policies that make (near) optimal reaction choices during each step of retrosynthetic planning. Using simulated experience or self-play, we train neural networks to estimate the expected synthesis cost or value of any given molecule based on a representation of its molecular structure. We show that learned policies based on this value network outperform heuristic approaches in synthesizing unfamiliar molecules from available starting materials using the fewest number of reactions. We discuss how the learned policies described here can be incorporated into existing synthesis planning tools and how they can be adapted to changes in the synthesis cost objective or material availability.

\end{abstract}

%%%%%%%%%%%%%%%%%%%%%%%%%%%%%%%%%%%%
%%%%%%%%%%%%%%%%%%%%%%%%%%%%%%%%%%%%
\section{Introduction}
\begin{figure}[p]
    \centering
    \includegraphics[width=\columnwidth]{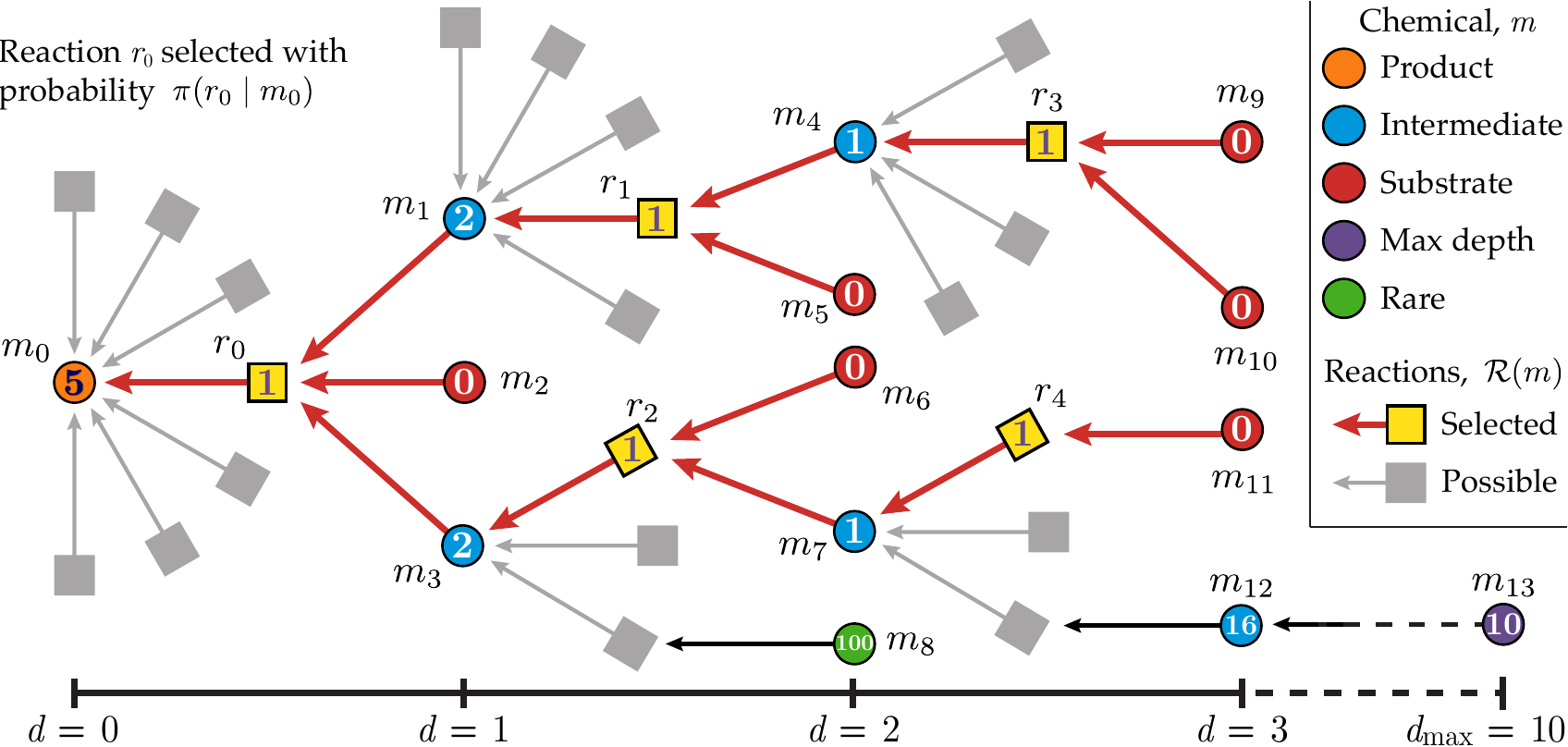}
    \caption{The objective of the retrosynthesis game is to synthesize the target product $m_0$ from available substrates by way of a synthesis tree that minimizes the cost function. Molecules and reactions are illustrated by circles and squares, respectively. Starting from the target, a reaction $r_0\in\mathcal{R}(m_0)$ (yellow) is selected according to a policy $\pi(r_0 | m_0)$ that links $m_0$ with precursors $m_1,m_2,m_3$. The grey squares leading to $m_0$ illustrate the other potential reactions in $\mathcal{R}(m_0)\setminus r_0$. The game continues one move at time reducing intermediate molecules (blue) until there are only substrates remaining, or until a maximum depth of 10 is reached. Rare molecules (green), which are not makeable under these constraints nor commercially available, are assigned a cost penalty of 100, while molecules at maximum depth (purple) are assigned a cost penalty of 10. Commercially available substrates are assigned zero cost. The synthesis cost of the product may be computed according to Eq.\ \ref{cost_equation} only on completion of the game. Here, the sampled pathway leading to the target (red arrows) has a cost of 5. }.
    \label{fig:1}
\end{figure}

The primary goal of computer-aided synthesis planning (CASP) is to help chemists accelerate the synthesis of desired molecules.\cite{warr2014short,engkvist2018computational,coley2018machine} Generally, a CASP program takes as input the structure of a target molecule and returns a sequence of feasible reactions linking the target to commercially available starting materials. The number of possible synthesis plans is often astronomical, and it is therefore desirable to identify the plan(s) that minimize some user-specified objective function (e.g., synthesis cost $C$). The challenge of identifying these optimal syntheses can be framed as a one-player game---the retrosynthesis game---to allow for useful analogies with chess and Go, for which powerful solutions based on deep reinforcement learning now exist \cite{silver2017mastering,silver_general_2018}. During play, the chemist starts from the target molecule and identifies a set of candidate reactions by which to make the target in one step (Figure~\ref{fig:1}). At this point, the chemist must decide which reaction to choose.  As in other games such as chess, the benefits of a particular decision may not be immediately obvious.  Only when the game is won or lost can one fairly assess the value of decisions that contributed to the outcome.  Once a reaction is selected, its reactant(s) become the new target(s) of successive retrosynthetic analyses. This branching recursive process of identifying candidate reactions and deciding which to use continues until the growing synthesis tree reaches the available substrates (a ``win'') or it exceeds a specified number of synthetic steps (a ``loss''). 

Winning outcomes are further distinguished by the cost $C$ of the synthesis pathway identified---the lower the better. This synthesis cost is often ambiguous and difficult to evaluate as it involves a variety of uncertain or unknown quantities. For example, the synthesis cost might include the price of the starting materials, the number of synthetic steps, the yield of each step, the ease of product separation and purification, the amount of chemical waste generated, the safety or environmental hazards associated with the reactions and reagents, \emph{et cetera}. It is arguably more challenging to accurately evaluate the cost of a proposed synthesis than it is to generate candidate syntheses. It is therefore common to adopt simple objective functions that make use of the information available (e.g., the number of reactions but not their respective yields).\cite{kowalik2012parallel} We refer to the output of any such function as the cost of the synthesis; optimal synthesis plans correspond to those with minimal cost. 

%%%%%%%%%%%%%%%%%%%%%%%%%%%%%%%%%%%%
%%%%%%%%%%%%%%%%%%%%%%%%%%%%%%%%%%%%

An expert chemist excels at the retrosynthesis game for two reasons: (1) she can identify a large number of feasible reaction candidates at each step, and (2) she can select those candidates most likely to lead to winning syntheses. These abilities derive from the chemist's prior knowledge and her past experience in making molecules. In contrast to games with a fixed rule set like chess, the identification of feasible reaction candidates (i.e., the possible ``moves'') is nontrivial: there may be thousands of possible candidates at each step using known chemistries. To address this challenge, computational approaches have been developed to suggest candidate reactions using libraries of reaction templates prepared by expert chemists\cite{szymkuc2016computer} or derived from literature data.\cite{law2009route,christ2012mining,bogevig2015route} Armed with these ``rules'' of synthetic chemistry, a computer can, in principle, search the entire space of possible synthesis pathways and identify the optimal one.

In practice, however, an exhaustive search of possible synthesis trees is not computationally feasible or desirable owing to the exponential growth in the number of reactions with distance from the target.\cite{grzybowski2009wired,kowalik2012parallel}  Instead, search algorithms generate a subset of possible synthesis trees, which may or may not contain the optimal pathway(s).  For longer syntheses, the subset of pathways identified is an increasingly small fraction of the total available.  Thus, it is essential to bias retrosynthetic search algorithms towards those regions of synthesis space most likely to contain the optimal pathway.  In the game of retrosynthesis, the player requires a strong guiding model, or \emph{policy}, for selecting the reaction at each step that leads to the optimal synthetic pathway(s). 

Prior reports on retrosynthetic planning have explored a variety of policies for guiding the generation of candidate syntheses.\cite{bertz1981first, sheridan2014modeling, szymkuc2016computer,coley2018scscore} These programs select among possible reactions using heuristic scoring functions,\cite{weininger1988smiles,szymkuc2016computer} crowd-sourced accessibility scores,\cite{ertl2009estimation,sheridan2014modeling} analogy to precedent reactions,\cite{coley_computer-assisted_2017} or parametric models (e.g., neural networks) trained on literature precedents.\cite{segler2017neural,segler2018planning} In particular, the Syntaurus software\cite{szymkuc2016computer} allows for user-specified scoring functions that can describe common strategies used by expert chemists\cite{corey1991logic} (e.g., using symmetric disconnections to favor convergent syntheses). By contrast, Segler and Waller used literature reaction data to train a neural network that determines which reaction templates are most likely to be effective on a given molecule.\cite{segler2017neural} The ability to rank order candidate reactions (by any means) allows for guiding network search algorithms (e.g., Monte Carlo tree search\cite{segler2018planning}) to generate large numbers of possible synthesis plans.  The costs of these candidates can then be evaluated to identify the ``best'' syntheses, which are provided to the chemist (or perhaps a robotic synthesizer). 

Here, we describe a different approach to retrosynthetic planning based on reinforcement learning,\cite{Sutton2017} in which the computer learns to select those candidate reactions that lead ultimately to synthesis plans minimizing a user-specified cost function. Our approach is inspired by the recent success of deep reinforcement learning in mastering combinatorial games such as Go using experience generated by repeated self-play \cite{silver2017mastering,silver_general_2018}. In this way, DeepMind's AlphaGo Zero learned to estimate the value of any possible move from any state in the game, thereby capturing the title of world champion.\cite{mnih2015human,silver2017mastering} Similarly, by repeated plays of the retrosynthesis game, the computer can learn which candidate reactions are most likely to lead from a given molecule to available starting materials in an optimal fashion. This approach requires no prior knowledge of synthetic strategy beyond the ``rules'' governing single-step reactions encoded in a library of reaction templates. Starting from a random policy, the computer explores the synthetic space to generate estimates of the synthesis cost for any molecule. These estimates form the basis for improved policies that guide the discovery of synthesis plans with lower cost.  This iterative process of policy improvement converges in time to optimal policies that identify the ``best'' pathway in a single play of the retrosynthesis game. Importantly, we show that (near) optimal policies trained on the synthesis of $\roughly$100,000 diverse molecules generalize well to the synthesis of unfamiliar molecules. We discuss how the learned policies described here can be incorporated into existing synthesis planning tools and how they can be adapted to different cost functions that reflect the changing demands of organic synthesis. 

%%%%%%%%%%%%%%%%%%%%%%%%%%%%%%%%%%%%
%%%%%%%%%%%%%%%%%%%%%%%%%%%%%%%%%%%%
\section{Results and Discussion}

%%%%%%%%%%%%%%%%%%%%%%%%%%%%%%%%%%%%
\subsection{The Retrosynthesis Game}

We formulate the problem of retrosynthetic analysis as a game played by a synthetic chemist or a computer program.  At the start of the game, the player is given a target molecule $m$ to synthesize starting from a set of buyable molecules denoted $\mathcal{B}$.  For any such molecule, there exists a set of reactions, denoted $\mathcal{R}(m)$, where each reaction includes the molecule $m$ as a product.  From this set, the player chooses a particular reaction $r\in\mathcal{R}(m)$ according to a policy $\pi(r\mid m)$, which defines the probability of selecting that reaction for use in the synthesis. The cost $c_{\rxn}(r)$ of performing the chosen reaction is added to a running total, which ultimately determines the overall synthesis cost.  Having completed one step of the retrosynthesis game, the player considers the reactant(s) of the chosen reaction in turn.  If a reactant $m^\prime$ is included among the buyable substrates, $m^\prime\in\mathcal{B}$, then the cost of that molecule $c_{\sub}(m^\prime)$ is added to the running total. Otherwise, the reactant $m^\prime$ must be synthesized following the same procedure outlined above.  This recursive processes results in a synthesis tree whose root is the target molecule and whose leaves are buyable substrates. The total cost of the resulting synthesis is 
\begin{equation}
\label{cost_equation}
    c_{\tot} = \sum_{r} c_{\rxn}(r) + \sum_{m} c_{\sub}(m),
\end{equation}
where the respective sums are evaluated over all reactions $r$ and all leaf molecules $m$ included in the final synthesis tree.  This simple cost function neglects effects due to interactions between successive reactions (e.g., costs incurred in switching solvents); however, it has the useful property that the expected cost $v_{\pi}(m)$ of making any molecule $m$ in one step via reaction $r$ is directly related to the expected cost of the associated reactants $m^\prime$,
\begin{equation}
    v_{\pi}(m) = \sum_{r\in\mathcal{R}(m)} \pi(r\mid m) \left[ c_{\rxn}(r) + \sum_{m'(r)} v_{\pi}(m^\prime)\right]. 
\label{eq:value}
\end{equation}
This recursive function terminates at buyable molecules of known cost, for which $v_{\pi}(m)=c_{\sub}(m)$ independent of the policy. 

The function $v_{\pi}(m)$ denotes the expected cost or ``value'' of any molecule $m$ under a specified policy $\pi$.  By repeating the game many times starting from many target molecules, it is possible to learn a parametric representation of this function valid for most molecules.  Importantly, knowledge of the value function under one policy $\pi$ enables the creation of new and better policies $\pi'$ that reduce the expected cost of synthesizing a molecule.  Using methods of reinforcement learning, such iterative improvement schemes lead to the identification of optimal policies $\pi_*$, which identify synthesis trees of minimal cost. The value of a molecule under such a policy is equal to the expected cost of selecting the ``best'' reaction at each step such that
\begin{equation}
    v_*(m) = \min_{r} \left[ c_{\rxn}(r) + \sum_{m'(r)} v_{*}(m^\prime)\right].
\end{equation}
From a practical perspective, the optimal value function takes as input a molecule (e.g., a representation of its molecular structure) and outputs a numeric value corresponding to the minimal cost with which it can be synthesized. 

Here, we considered a set of 100,000 target molecules selected from the Reaxys database on the basis of their structural diversity (see Methods). The set of buyable substrates $\mathcal{B}$ contained $\roughly$300,000 molecules selected from the Sigma-Aldrich,\cite{sigma-aldrich} eMolecules,\cite{emoles} and LabNetwork\cite{labnet} catalogs that have list prices less than \$100/g. At each step, the possible reactions $\mathcal{R}(m)$ were identified using a set of 60,000 reaction templates derived from more than 12 million single-step reaction examples reported in the Reaxys database (see Methods). Because the application of reaction templates is computationally expensive, we used a template prioritizer to identify those templates most relevant to a given molecule.\cite{segler2017neural} On average, this procedure resulted in up to 50 possible reactions for each molecule encountered during synthesis planning. We assume the space of molecules and reactions implicit in these transformations is representative of real organic chemistry while recognizing the inevitable limitations of templates culled from incomplete and sometimes inaccurate reaction databases.  For simplicity, the cost of each reaction step was set to one, $c_{\rxn}(r)=1$, and the substrate costs to zero, $c_{\sub}(m)=0$. With these assignments, the cost of making a molecule is equivalent to the number of reactions in the final synthesis tree.

To prohibit the formation of unreasonably deep synthesis trees, we limited our retrosynthetic searches to a maximum depth of $d_{\max}=10$.  As detailed in the Methods, the addition of this termination criterion to the recursive definition of the value function (\ref{eq:value}) requires some minor modifications to the retrosynthesis game.  In particular, the expected cost of synthesizing a molecule $m$ depends also on the residual depth, $v_{\pi}= v_{\pi}(m,\delta)$, where $\delta=d_{\max} - d$ is the difference between the maximum depth and the current depth $d$ within the tree. If a molecule $m$ not included among the buyable substrates is encountered at a residual depth of zero, it is assigned a large cost $v_{\pi}(m,0)=P_1$, thereby penalizing the failed search. Additionally, in the event that no reactions are identified for a given molecule $m$ ($\mathcal{R}(m)=\varnothing$), we assign an even larger penalty $P_2$, which encourages the player to avoid such molecules if possible.  Below, we use the specific numeric penalties of $P_1=10$ and $P_2=100$ for all games.

%%%%%%%%%%%%%%%%%%%%%%%%%%%%%%%%%%%%
\subsection{Heuristic Policies}

En route to the development of optimal policies for retrosynthesis, we first consider the performance of some heuristic policies that provide context for the results below.  Arguably the simplest policy is one of complete ignorance, in which the player selects a reaction at random at each stage of the synthesis---that is, $\pi(r\mid m)=\text{constant}$. We use this ``straw man'' policy to describe the general process of policy evaluation and provide a baseline from which to measure subsequent improvements.   

During the evaluation process, the computer plays the retrosynthesis game to the end making random moves at each step of the way.  After each game, the cost of each molecule in the resulting synthesis tree is computed. This process is repeated for each of the 100,000 target molecules considered. These data points---each containing a molecule $m$ at residual depth $\delta$ with cost $c$---are used to update the parametric approximation of the value function $v_{\pi}(m,\delta)$.  As detailed in the Methods, the value function is approximated by a neural network that takes as input an extended-connectivity (ECFP) fingerprint of the molecule $m$ and the residual depth $\delta$ and outputs a real valued estimate of the expected cost under the policy $\pi$.\cite{rogers2010extended} This process is repeated in an iterative manner as the value estimates of the target molecules, $v_{\pi}(m,d_{\max})$, approach their asymptotic values.  

\begin{figure}[p]
    \centering
    \includegraphics[width=3.5in]{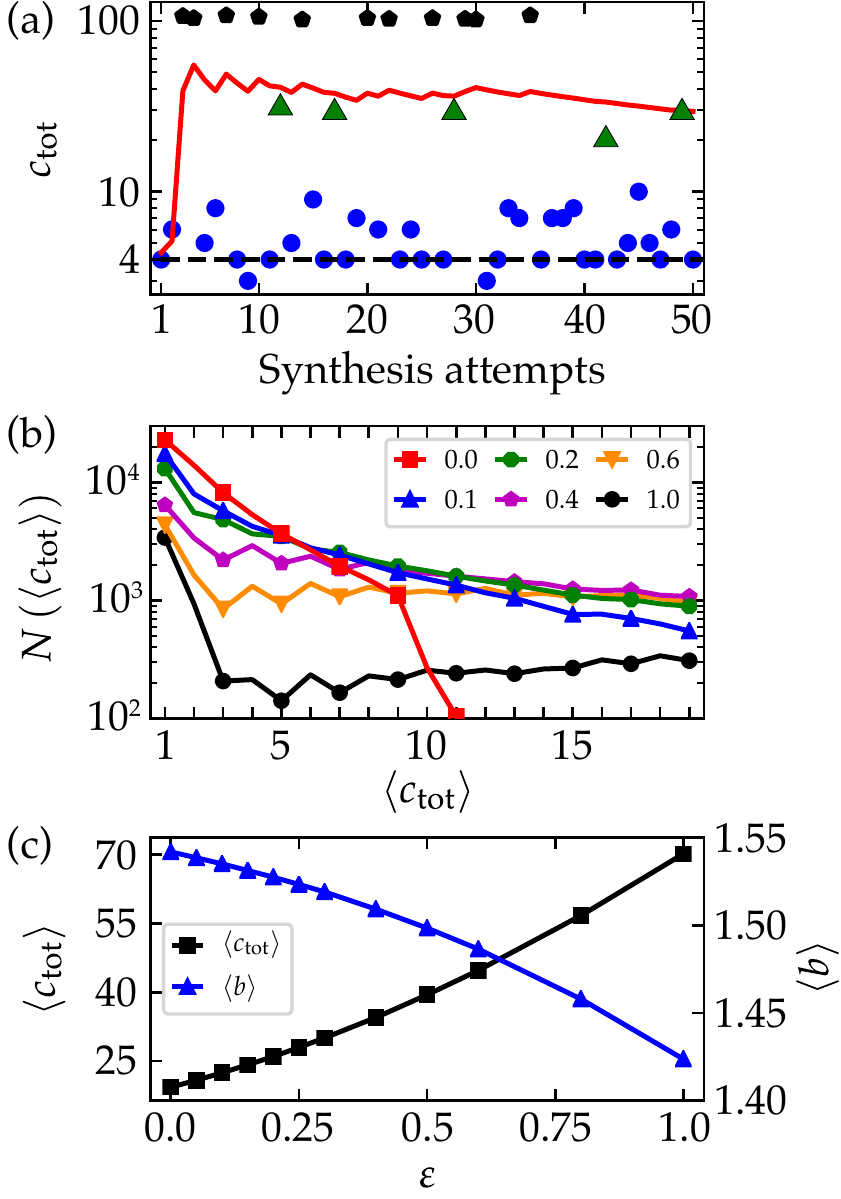}
    \caption{Heuristic policies. (a) Synthesis cost $c_{\tot}$ for a single molecule $m$ (N-dibutyl-4-acetylbenzeneacetamide) for successive iterations of the retrosynthesis game under the random policy. Blue circles denote ``winning'' synthesis plans that trace back to buyable molecules. Green triangles and black pentagons denote ``losing'' plans that exceed the maximum depth or include unmakable molecules, respectively. The solid line shows the neural network prediction of the value function $v_{\pi}(m,d_{\max})$ as it converges to the average synthesis cost. The dashed line shows the expected cost under the deterministic ``symmetric disconnection'' policy with $\gamma = 1.5$. (b) Distribution of expected costs $v_{\pi}(m,d_{\max})$ over the set of 100,000 target molecules for different noise levels $\varepsilon$. The red squares and black circles show the performance of the symmetric disconnection policy ($\varepsilon=0$) and the random policy ($\varepsilon=1$), respectively. (c) The average synthesis cost of the target molecules decreases with increasing noise level $\varepsilon$, while the average branching factor increases. Averages were estimated from 50 plays for each target molecule.}
    \label{fig:2}
\end{figure}

Figure~\ref{fig:2}a shows the total synthesis cost $c_{\tot}$ for a single target molecule under the random policy (markers). Each play of the retrosynthesis game has one of three possible outcomes: a ``winning'' synthesis plan terminating in buyable substrates (blue circles), a ``losing'' plan that exceeds the maximum depth (green triangles), and a ``losing'' plan that contains dead-end molecules that cannot be bought or made (black pentagons).  After many synthesis attempts, the running average of the fluctuating synthesis cost converges to the expected cost $v_{\pi}(m,d_{\max})$ as approximated by the neural network (red line).  Repeating this analysis for the 100,000 target molecules, the random policy results in an average cost of $\roughly$110 per molecule with only a 25\% chance of identifying a winning synthesis in each attempt. Clearly, there is room for improvement.

%%%%%%%%%%%%%%%%%%%%%%%%%%%%%%%%%%%%
%%%%%%%%%%%%%%%%%%%%%%%%%%%%%%%%%%%%

Beyond the random policy, even simple heuristics can be used to improve performance significantly.  In one such policy, inspired by Syntaurus,\cite{szymkuc2016computer} the player selects the reaction $r$ that maximizes the quantity
\begin{equation}
    f(r) = n_s(m)^\gamma - \sum_{m'(r)} n_s(m^\prime)^\gamma
\label{heuristic}
\end{equation}
where $n_s(m)$ is the length of the canonical smiles string representing molecule $m$, $\gamma$ is a user-specified exponent, and the sum is taken over all reactants $m^\prime$ associated with a reaction $r$.  When $\gamma>1$, the reactions that maximize this function can be interpreted as those that decompose the product into multiple parts of roughly equal size.  Note that in contrast to the random policy, this greedy heuristic is deterministic: each play of the game results in the same outcome. Figure~\ref{fig:2}a shows the performance of this ``symmetric disconnection'' policy with $\gamma=1.5$ for a single target molecule (dashed line).  Interestingly, while the pathway identified by the greedy policy is much better on average than those of the random policy ($c_{\tot} = 4$ vs.~$\langle c_{\tot}\rangle = 35.1$), repeated application of the latter reveals the existence of an even better pathway containing only three reactions. An optimal policy would allow for the identification of that best synthesis plan during a \emph{single} play of the retrosynthesis game.

The performance of a policy is characterized by the distribution of expected costs over the set of target molecules. Figure~\ref{fig:2}b shows the cost distribution for a series of policies that interpolate between the greedy ``symmetric disconnection'' policy and the random policy (see also Figure~\ref{fig:Sa}). The intermediate $\varepsilon$-greedy policies behave greedily with probability $1-\varepsilon$, selecting the reaction that maximizes $f(r)$, but behave randomly with probability $\varepsilon$, selecting any one of the possible reactions $r\in\mathcal{R}(m)$ with equal probability.  On average, the addition of such noise is usually detrimental to policy performance. Noisy policies are less likely to identify a successful synthesis for a given target (Figure~\ref{fig:Sb}a) and result in longer syntheses when they do succeed (Figure~\ref{fig:Sb}b). Consequently, the average cost $\langle c_{\tot}\rangle$ increases monotonically with increasing noise as quantified by the parameter $\varepsilon$ (Figure~\ref{fig:2}c).

The superior performance (lower synthesis costs) of the greedy  policy is correlated with the average branching factor $\langle b \rangle$, which represents the average number of reactants for each reaction in the synthesis tree. Branching is largest for the greedy policy ($\varepsilon=0$) and decreases monotonically with increasing $\varepsilon$ (Figure~\ref{fig:2}c). On average, synthesis plans with greater branching (i.e., convergent syntheses) require fewer synthetic steps to connect the target molecules to the set of buyable substrates. This observation supports the chemical intuition underlying the symmetric disconnection policy: break apart each ``complex'' molecule into ``simpler'' precursors.  However, this greedy heuristic can sometimes be short-sighted. An optimal retrosynthetic ``move'' may increase molecular complexity in the short run to reach simpler precursors more quickly in the longer run (e.g., in protecting group chemistry). An optimal policy would enable the player to identify local moves (i.e., reactions) that lead to synthesis pathways with minimum total cost. 

%%%%%%%%%%%%%%%%%%%%%%%%%%%%%%%%%%%%
\subsection{Policy improvement through self-play}

Knowledge of the value function, $v_{\pi}$, under a given policy $\pi$ enables the identification of better policies that reduce the expected synthesis cost. To see this, consider a new policy $\pi'$ that selects at each step the reaction that minimizes the expected cost under the old policy $\pi$
\begin{equation}
    \pi'(r\mid m) = \begin{dcases} 1 & \text{if } r = \argmin_{r\in\mathcal{R}(m)} \left[ c_{\rxn}(r) + \sum_{m'(r)} v_{\pi}(m^\prime)  \right] \\ 0 & \text{otherwise} \end{dcases}.
\label{eq:greedy}
\end{equation}
By the policy improvement theorem\cite{Sutton2017}, this greedy policy $\pi'$ is guaranteed to be as good or better than the old policy $\pi$---that is, $v_{\pi'} \leq v_{\pi}$, where equality holds only for the optimal policy.  This result provides a basis for systematically improving any policy in an iterative procedure called policy iteration\cite{Sutton2017}, in which the value function $v_{\pi}$ leads to an improved policy $\pi'$ that leads to a new value function $v_{\pi'}$ and so on. 

One of the challenges in using the greedy policy Eq.\ \ref{eq:greedy} is that it generates only a single pathway and its associated cost for each of the target molecules. The limited exposure of these greedy searches can result in poor estimates of the new value function $v_{\pi'}$, in particular for molecules that are not included in the identified pathways. A better estimate of $v_{\pi'}$ can be achieved by exploring more of the molecule space in the neighborhood of these greedy pathways.  Here, we encourage exploration by using an $\varepsilon$-greedy policy, which introduces random choices with probability $\varepsilon$ but otherwise follows the greedy policy Eq.\ \ref{eq:greedy}.  Iteration of this $\varepsilon$-soft policy is guaranteed to converge to an optimal policy that minimizes the expected synthesis cost for a given noise level $\varepsilon > 0$.\cite{Sutton2017}  Moreover, by gradually lowering the noise level, it is possible to approach the optimal greedy policy in the limit as $\varepsilon\rightarrow0$.

%%%%%%%%%%%%%%%%%%%%%%%%%%%%%%%%%%%%
\subsection{Training protocol}

Starting from the random policy, we used self-play to learn an improved policy over the course of 1000 iterations, each comprised of $\roughly$100,000 retrosynthesis games initiated from the target molecules. During the first iteration, each target molecule was considered in turn using the $\varepsilon$-greedy policy Eq.\ \ref{eq:greedy} with $\varepsilon=0.2$. Candidate reactions and their associated reactants were identified by application of reaction templates as detailed in the Methods. Absent an initial model of the value function, the expected costs of molecules encountered during play were selected at random from a uniform distribution on the interval $\left[1,100\right]$. Following the completion of each game, the costs of molecules in the selected pathway were computed and stored for later use. In subsequent iterations, the values of molecules encountered previously (at a particular depth) were estimated by their average cost. After the first 50 iterations, the value estimates accumulated during play were used to train a neural network, which allowed for estimating the values of new molecules not encountered during the previous games (see Methods for details on the network architecture and training). Policy improvement continued in an iterative fashion guided both by the average costs (for molecules previously encountered) and by the neural network (for new molecules), which was updated every 50-100 iterations.

During policy iteration, the noise parameter was reduced from $\varepsilon=0.2$ to $0$ in increments of 0.05 every 200 iterations in an effort to anneal the system towards an optimal policy. Following each change in $\varepsilon$, the saved costs were discarded such that subsequent value estimates were generated at the current noise level $\varepsilon$. The result of this training procedure was a neural network approximation of the (near) optimal value function $v_{*}(m,\delta)$, which estimates the minimum cost of synthesizing any molecule $m$ starting from residual depth $\delta$. In practice, we found that a slightly better value function could be obtained using the cumulative reaction network generated during policy iteration. Following Kowalik \emph{et al.},\cite{kowalik2012parallel} we used dynamic programming to compute the minimum synthesis cost for each molecule in the reaction network. These minimum costs were then used to train the final neural network approximation of the value function $v_{*}$.  

%%%%%%%%%%%%%%%%%%%%%%%%%%%%%%%%%%%%
\begin{figure}[t]
\centering
\includegraphics[width=3.5in]{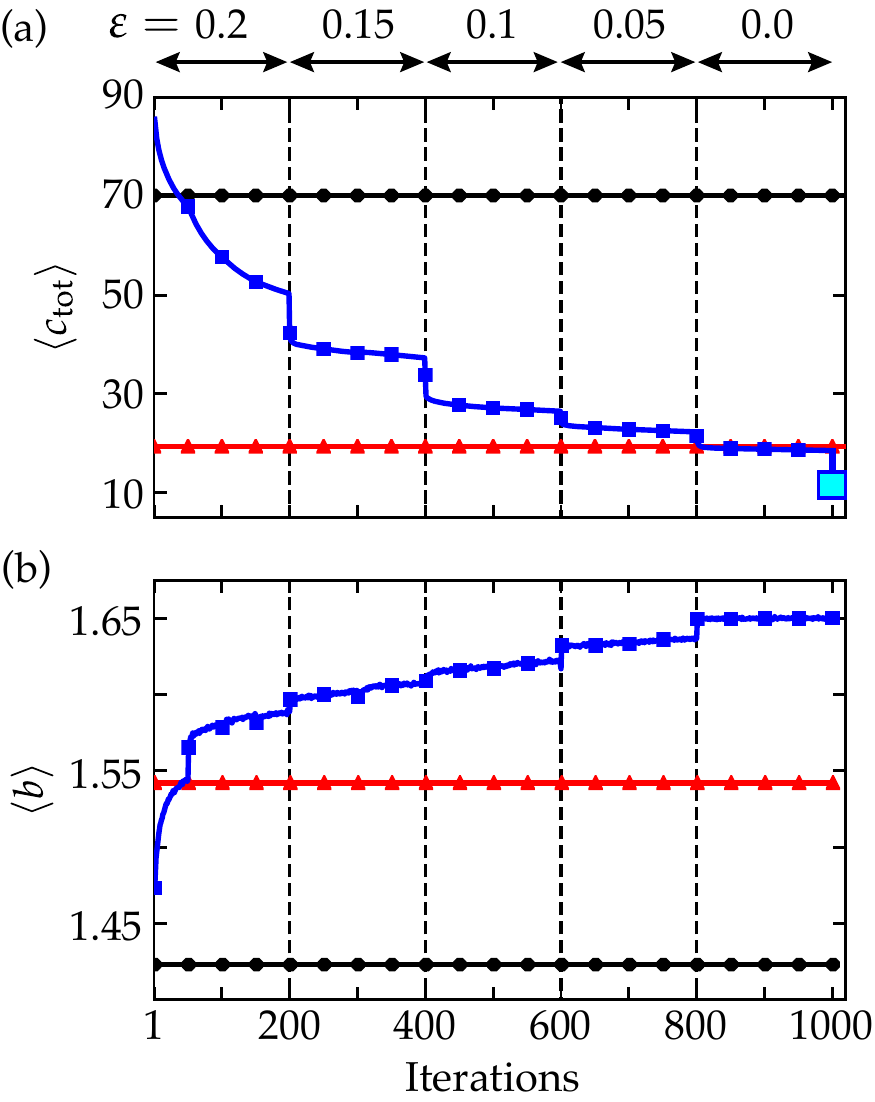}
\caption{Training results. In (a) and (b) $\langle c_\text{tot} \rangle$ and $\langle b_\text{tot} \rangle$ computed using $\pi^*$ are plotted versus policy iterations, respectively (solid blue squares).  Solid horizontal lines show these quantities for the heuristic policy $\pi_{sd}$ (red triangles) and the random policy (black circles).  The larger cyan square shows $\langle c_\text{tot} \rangle$ after each tree had been searched for the best (lowest) target cost. Dashed vertical lines show points when $\varepsilon$ was lowered.} 
\label{fig4}
\end{figure}
%%%%%%%%%%%%%%%%%%%%%%%%%%%%%%%%%%%%

%%%%%%%%%%%%%%%%%%%%%%%%%%%%%%%%%%%%
\subsection{Training results}

Figure \ref{fig4}a shows how the average synthesis cost $\langle c_\text{tot} \rangle$ decreased with each iteration over the course of the training process. Initially, the average cost was similar to that of the random policy ($\langle c_\text{tot} \rangle\approx 70$) but improved steadily as the computer learned to identify ``winning'' reactions that lead quickly to buyable substrates. After 800 iterations, the cost dropped below that of the symmetric disconnection policy ($\langle c_\text{tot} \rangle = 19.3$) but showed little further improvement in the absence of exploration (i.e., with $\varepsilon=0$). The final cost estimate ($\langle c_\text{tot} \rangle = 11.4$, cyan square) was generated by identifying the minimum cost pathways present in the cumulative reaction network generated during the training process. The final drop in cost for $\varepsilon=0$ suggests that further policy improvements are possible using improved annealing schedules. We emphasize that the final near-optimal policy was trained from a state of complete ignorance, as directed by the user-specified objective function to minimize the synthesis cost.

During the training process, the decrease in synthesis cost was guided both by motivation, as prescribed by the cost function, and by opportunity, as dictated by the availability of alternate pathways. Early improvements in the average cost were achieved by avoiding ``unmakable'' molecules, which contributed the largest cost penalty, $P_2=100$. Of the target molecules, 11\% reduced their synthesis cost from $c_\text{tot} > P_2$ to $P_2 > c_\text{tot} > P_1$ by avoiding such problematic molecules. By contrast, only 2\% of targets improved their cost from $P_2 > c_\text{tot} > P_1$ to $P_1 > c_\text{tot}$. In other words, if a synthesis tree was not found initially at a maximum depth of $d_{\max}=10$, it was unlikely to be discovered during the course of training. Perhaps more interesting are those molecules (ca.\ ~10\%) for which syntheses were more easily found but subsequently improved (i.e., shortened) during the course of the training process.  See Table~\ref{table_stats} for a more detailed breakdown of these different groups.

%%%%%%%%%%%%%%%%%%%%%%%%%%%%%%%%%%%%
\begin{table}[t]
\renewcommand{\arraystretch}{1.3}
\centering
\begin{tabular}{ ccc|cc } 

    \toprule

    \multicolumn{1}{c}{} & \multicolumn{2}{c|}{\textbf{Train} (100,00)} & \multicolumn{2}{c}{\textbf{Test} (25,000)} \\

    & $\pi_{sd}$ & $\pi_*$ & $\pi_{sd}$ & $\pi_*$ \\
 
    \midrule
    $\langle c_\text{tot} \rangle$ & 19.3 & 13.1 & 19.2 & 11.5 \\
    $\langle b \rangle$ & 1.54 & 1.65 & 1.54 & 1.58 \\
    $c_\text{tot} < P_1$ & 64\pct & 83\pct & 65\pct & 73\pct \\
    $P_1 \le c_\text{tot} < P_2$ & 25\pct & 11\pct & 24\pct & 22\pct \\
    $c_\text{tot} \ge P_2$ & 11\pct & 6\pct & 11\pct & 5\pct \\ 
    \bottomrule
\end{tabular}
\caption{\label{table_stats} Training and testing results for the symmetric disconnection policy $\pi_{sd}$ and the learned policy $\pi_*$. Percentages were computed based the sizes of the training set ($\roughly$100,000) and the testing set ($\roughly$25,000).} %Actual numbers - 95,774, 23,945. 
\end{table}
%%%%%%%%%%%%%%%%%%%%%%%%%%%%%%%%%%%%

Consistent with our observations above, lower cost pathways were again correlated with the degree of branching $b$ along the synthesis trees (Figure~\ref{fig4}b). Interestingly, the average branching factor for synthesis plans identified by the learned policy was significantly larger than that of the symmetric disconnection policy ($\langle b\rangle = 1.65$ vs.\ $1.54$). While the latter favors branching, it does so locally based on limited information---namely, the heuristic score of Eq.\ \ref{eq:greedy}. By contrast, the learned policy uses information provided in the molecular fingerprint to select reactions that increase branching across the entire synthesis tree (not just the single step). Furthermore, while the heuristic policy favors branching \emph{a priori}, the learned policy does so only in the service of reducing the total cost. Changes in the objective function (e.g., in the cost and availability of the buyable substrates) will lead to different learned policies.

%%%%%%%%%%%%%%%%%%%%%%%%%%%%%%%%%%%% 
\subsection{Model validation}

%%%%%%%%%%%%%%%%%%%%%%%%%%%%%%%%%%%%
\begin{figure}[t]
\centering
\includegraphics[width=6.6in]{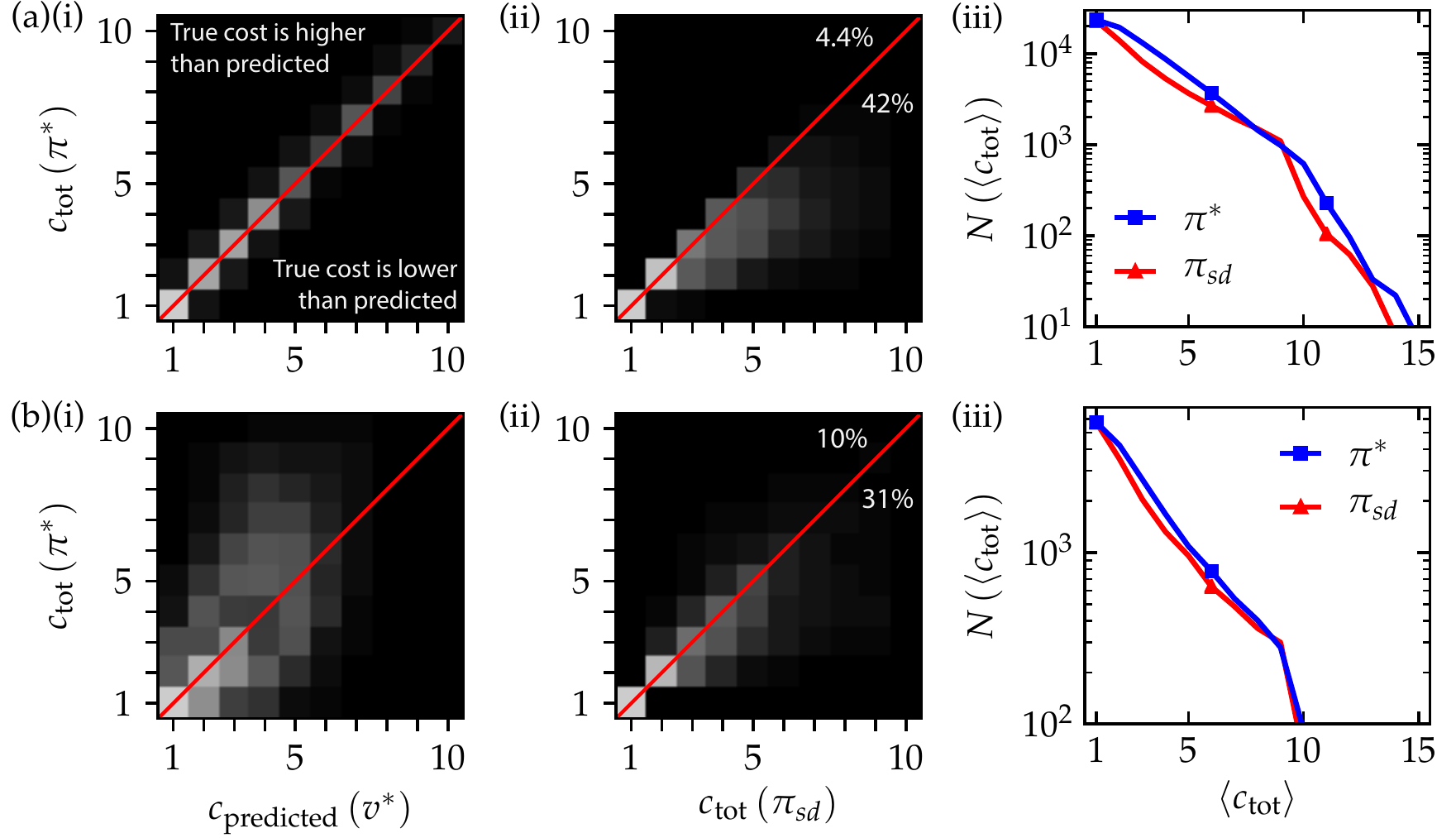}
\caption{Model Validation. In (a) and (b) a 2D histogram illustrates the relationship between the synthesis cost $c_{\tot}$ determined by the learned policy $\pi_*$ and that predicted by the value network $v_*$ for each of the $\roughly$100,000 training molecules and the $\roughly$25,000 testing molecules, respectively. In (c) and (d), a 2D histogram compares the synthesis cost $c_{\tot}$ determined by the symmetric disconnection policy $\pi_{sd}$ to that of learned policy $\pi_*$ for training and testing molecules, respectively. In (a-d), the gray-scale intensity is linearly proportional to the number of molecules within a given bin; the red line shows the identity relation. In (c) and (d), the percentage of molecules for which $\pi_*$ ($\pi_{sd}$) found the cheaper pathway is listed below (above) the red line.  In (e) and (f), distributions of synthesis costs $c_{\tot}$ determined under policies $\pi_{sd}$ and $\pi_*$ are shown for both testing and testing molecules, respectively.}
\label{fig5}
\end{figure}
%%%%%%%%%%%%%%%%%%%%%%%%%%%%%%%%%%%%

Figure~\ref{fig5} compares the performance of the learned policy evaluated on the entire set of $\roughly$100,000 target molecules used for training and on a different set of $\roughly$25,000 target molecules set aside for testing. For the training molecules, the value estimates $v_*(m)$ predicted by the neural network are highly correlated with the actual costs obtained by the final learned policy $\pi_*$ (Figure~\ref{fig5}a). We used the same near-optimal policy to determine the synthesis cost of the testing molecules, $c_{\tot}(\pi_*)$. As illustrated in Figure~\ref{fig5}b, these costs were correlated to the predictions of the value network $v_*(m)$ albeit more weakly than those of the training data (Pearson coefficient of 0.5 for testing vs. 0.99 for training). This correlation was stronger for the data in Figure~\ref{fig5}b, which focuses on those molecules that could actually be synthesized (Pearson coefficient of 0.7 for the 73\% testing molecules with ``winning'' syntheses).  

Figure~\ref{fig5}c,d compares the synthesis costs of the symmetric disconnection policy $\pi_{sd}$ against that of the learned policy $\pi_*$ for both the training and testing molecules. The figure shows that the results are highly correlated (Pearson coefficient 0.84 and 0.86 for training and testing, respectively), indicating that the two policies make similar predictions. However, closer inspection reveals that the learned policy is systematically better than the heuristic as evidenced by the portion of the histogram below the diagonal (red line). For these molecules (42\% and 31\% of the training and testing sets, respectively), the learned policy identifies synthesis trees containing fewer reactions than those of the heuristic policy during single deterministic plays of the retrosynthesis game. By contrast, it is rare in both the training and testing molecules (about 4\% and 11\%, respectively) that the symmetric disconnection policy performs better than the learned policy.  Additionally, the learned policy is more likely to succeed in identifying a viable synthesis plan leading to buyable substrates (Figure~\ref{fig5}c). Of the $\roughly$25,000 testing molecules, ``winning'' synthesis plans were identified for 73\% using the learn policy as compared to 64\% using the heuristic. These results suggest that the lessons gleaned from the training molecules can be used to improve the synthesis of new and unfamiliar molecules.

%%%%%%%%%%%%%%%%%%%%%%%%%%%%%%%%%%%%
\section{Conclusions}

We have shown that reinforcement learning can be used to identify effective policies for the computational design of retrosynthetic pathways. In this approach, one specifies the global objective function to be minimized (here, the synthesis cost) without the need for \emph{ad hoc} models or heuristics to guide local decisions during generation of the synthesis plan. Starting from a random policy, repeated plays of the retrosynthesis game are used to systematically improve performance in an iterative process that converges in time to an optimal policy. The learned value function provides a convenient estimate for the synthesis cost of any molecule, while the associated policy allows for rapid identification of the synthesis path.  Importantly, the cost function can be easily adapted, for example to include specific costs for reactions and/or buyable substrates.  Policy iteration using a different cost function will result in a different policy that reflects the newly specified objectives.

The chemical feasibility of synthetic pathways identified by the learned policy are largely determined by the quality of the reaction templates. The present templates are derived algorithmically from reaction precedents reported in the literature; however, an identical approach based on reinforcement learning could be applied using template libraries curated by human experts \cite{szymkuc2016computer, klucznik2018efficient}. Alternatively, it may be possible to forgo the use of reaction templates altogether in favor of machine learning approaches that suggest reaction precursors by other means.\cite{liu_retrosynthetic_2017}  Ideally, such predictions should be accompanied by recommendations regarding the desired conditions for performing each reaction in high yield.\cite{marcou2015expert, lin2016automatized, segler2017modelling,gao_using_2018}

In the present approach, the deterministic policy learned during training is applied only once to suggest one (near) optimal synthesis pathway. Additional pathways are readily generated, for example using Monte-Carlo Tree Search (MCTS) to bias subsequent searches away from previously identified pathways.\cite{segler2017neural} A similar approach is used by Syntaurus, which relies on heuristic scoring functions to guide the generation of many possible synthesis plans, from which the ``best’’ are selected. The main advantage of a strong learned policy is to direct such exploration more effectively towards these best syntheses, thereby reducing the computational cost of exploration. 

We note, however, that the computational costs of training the learned policy are significant (ca.\ ~several million CPU hours for the training in Figure~\ref{fig4}). While the application of reaction templates remains the primary bottleneck (ca.\ $\roughly$50\%), the additional costs of computing ECFP fingerprints and evaluating the neural network were a close second (ca.\ $\roughly$45\%). These costs can be greatly reduced by using simple heuristics to generate synthetic pathways, from which stronger policies can be learned. We found that Eq.\ \ref{heuristic} performed remarkably well and was much faster to evaluate than the neural network.  Such fast heuristics could be used as starting points for iterative policy improvement or as roll-out policies within MCTS-based learning algorithms.\cite{Sutton2017}  This approach is conceptually similar to the first iteration of AlphaGo introduced by DeepMind.\cite{silver2016mastering}. Looking forward, we anticipate that the retrosynthesis game will soon follow the way of chess and go, in which self-taught algorithms consistently outperform human experts.

%%%%%%%%%%%%%%%%%%%%%%%%%%%%%%%%%%%%
\section{Methods}

%%%%%%%%%%%%%%%%%%%%%%%%%%%%%%%%%%%%
\subsection{Target molecules}

Training/testing sets of 95,774/23,945 molecules were selected from the Reaxys database on the basis of their structural diversity. Starting from more than 20 million molecules in the database, we excluded (i) those listed in the database of buyable compounds, (ii) those with SMILES strings shorter than 20 or longer than 100, and (iii) those with multiple fragments (i.e., molecules with `.' in the SMILES string). The resulting $\roughly16$ million molecules were then aggregated using the Taylor-Butina (TB) algorithm\cite{butina1999unsupervised} to form $\roughly1$ million clusters, each comprised of ``similar'' molecules. Structural similarity between two molecules $i$ and $j$ was determined by the Tanimoto coefficient
\begin{equation}
    T = \frac{m_i \cdot m_j}{|m_i|^2 + |m_j|^2 - |m_i \cdot m_j|},
\end{equation}
where $m_i$ is the ECFP4 fingerprint for molecule $i$.\cite{glen2006circular} We used fingerprints of length 1024 and radius 3. Two molecules within a common cluster were required to have a Tanimoto coefficient of $T > 0.4$. The target molecules were chosen as the centroids of the $\roughly125,000$ largest clusters, each containing more than 20 molecules and together representing more than $\roughly12$ million molecules. These target molecules were partitioned at random to form the final sets for training and testing.

%%%%%%%%%%%%%%%%%%%%%%%%%%%%%%%%%%%%
\subsection{Buyable molecules}

A molecule is defined to be a substrate if it is listed in the commercially available Sigma-Aldrich,\cite{sigma-aldrich} eMolecules,\cite{emoles} or LabNetwork catalogs,\cite{labnet} and does not cost more than \$100/g. The complete set of molecules in these catalogs with $\text{price per gram} \le \$100$ is denoted $\mathcal{B} = \{m_1,\dots, m_n\}$ with $n \approx 300,000$. The molecules contained in each catalog can be downloaded by visiting the respective webpage.

%%%%%%%%%%%%%%%%%%%%%%%%%%%%%%%%%%%%
\subsection{Reaction templates}

Given a molecule $m$, we used a set of $\roughly60,000$ reaction templates to generate sets of possible precursors $m^\prime$, which can be used to synthesize $m$ in one step.  As detailed previously,\cite{coley2017prediction} these templates were extracted automatically from literature precedents and encoded using the SMARTS language. The application of the templates involves two main steps, substructure matching and bond rewiring, which were implemented using RDkit\cite{rdkit}. Briefly, we first search the molecule $m$ for a structural pattern specified by the template. For each match, the reaction template further specifies the breaking and making of bonds among the constituent atoms to produce the precursor molecule(s) $m^\prime$. We used the RDChiral package\cite{rdchiral} to handle the creation, destruction, and preservation of chiral centers during the reaction.

The application of reaction templates to produce candidate reactions represents a major computational bottleneck in the retrosynthesis game due to the combinatorial complexity of substructure matching.  Additionally, even when a template generates a successful match, it may fail to account for the larger molecular context resulting in undesired byproducts during the forward reaction. These two challenges can be partially alleviated by use of a ``template prioritizer'',\cite{segler2017neural} which takes as input a representation of the target molecule $m$ and generates a probability distribution over the set of templates based on their likelihood of success. By focusing only on the most probable templates, the prioritizer can serve to improve both quality of the suggested reactions and the speed with which they are generated. In practice, we trained a neural network prioritizer on 5.4 million reaction examples from Reaxys and selected  the top 99.5\% of templates for each molecule $m$ encountered. This filtering process drastically reduced the total number templates applied from 60,000 to less than 50 for most molecules. The training and validation details as well as the model architecture are available on Github\cite{retrotemp}. 
%The prioritizer model used here (5d4M) predicts reasonable accuracies after training on 5.4 million examples. For example, $T(s)$ yields a top-1 accuracy of 43.1\%, top-10 of 78.3\%, top-50 of 89.8\%, and top-100 of 92.6\%.

%%%%%%%%%%%%%%%%%%%%%%%%%%%%%%%%%%%%
\subsection{Policy Iteration}

As noted in the main text, the depth constraint imposed on synthesis trees generated during the retrosynthesis requires some minor modifications to the value function of Eq.\ \ref{eq:value}.  The expected cost of synthesizing a molecule $m$ now depends on the residual depth $\delta$ as 
\begin{equation}
    v_{\pi}(m, 
    \delta) = \sum_{r\in\mathcal{R}(m)} \pi(r\mid m,\delta) \left[ c_{\rxn}(r) + \sum_{m'(r)} v_{\pi}(m^\prime,\delta-1)\right],
\end{equation}
where the first sum is over candidate reactions with $m$ as product, and the second is over all reactants $m^\prime(r)$ associated with a reaction $r$. For the present cost model, the expected cost $ v_{\pi}(m^\prime,\delta)$ increases with decreasing $\delta$ due to the increased likelihood of being penalized (to the extent $P_1$) for reaching the maximum depth ($d=d_{\max}$ such that $\delta=0$). Similarly, the $\varepsilon$-greedy policy used in policy improvement must also account for the residual depth at which a molecule is encountered
\begin{equation}
    \pi(r\mid m,\delta) = \begin{dcases} 1-\varepsilon & \text{if } r = \argmin_{r\in\mathcal{R}(m)} \left[ c_{\rxn}(r) + \sum_{m'(r)} v_{\pi}(m^\prime,\delta-1)  \right] \\ \varepsilon & \text{otherwise} \end{dcases}.
\end{equation}
These recursive functions are fully specified by three terminating conditions introduced in the main text: (1) buyable molecule encountered, $v(m,\delta\neq0)=c_{\sub}(m)$ for $m\in\mathcal{B}$; (2) maximum depth reached, $v(m,0)=P_1$; and (3) unmakeable molecule encountered, $v(m,\delta\neq0)=P_2$ for $\mathcal{R}(m)=\varnothing$.

%%%%%%%%%%%%%%%%%%%%%%%%%%%%%%%%%%%%
\subsection{Neural network architecture and training}

\begin{figure} [h]
\centering
\includegraphics{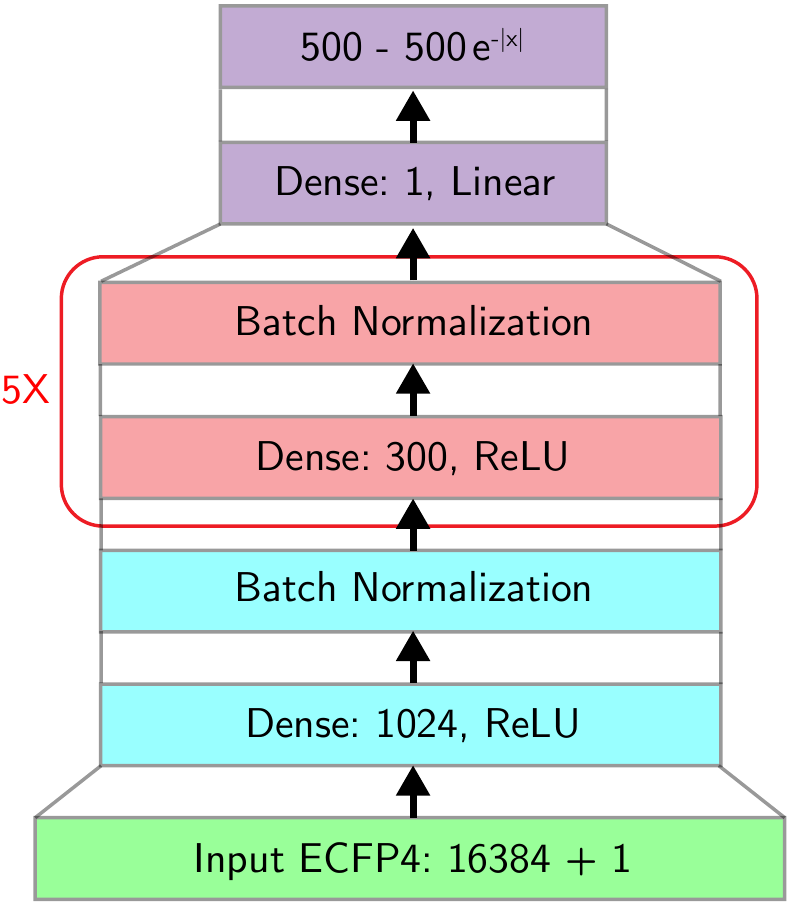}
\caption{The neural model for the cost of molecules is a feed-forward neural network that accepts as input (green) an ECFP fingerprint of size 16384 extended to include the residual depth $\delta$ of the molecule. The architecture includes one input later (blue) consisting of 1024 nodes, five hidden layers (red) each containing 300 nodes, and one output layer (purple) of size one plus a filter (also purple) that scales the initial output number to be within the range $\left[0,500\right]$. We also used batch normalization after each layer. The final output represents the estimated cost.}
\label{fig:model_arch}
\end{figure}

We employed a multi-layer neural network illustrated schematically in Figure~\ref{fig:model_arch}. The 17 million model parameters were learned using gradient descent on training data generated by repeated plays of the retrosynthesis game. Training was performed using Keras with the Theano backend and the Adam optimizer with an initial learning rate of $0.001$, which decayed with the number of model updates $k$ as  $0.001 / (1 + 2\sqrt{k})$ (13 updates were used to compute $\pi_*$). During each update, batches of 128 molecules and their computed average costs at a fixed $\varepsilon$ were selected from the most recent data and added to a replay buffer. Batches of equivalent size were randomly selected from the buffer and passed through the model for up to 100 epochs (1 epoch was taken as the total number of new data points having passed through the network). The mean-average error between the averaged (true) and predicted costs was used as the loss function. The latest model weights were then used as the policy for the next round of synthesis games. 
%, which consists of the molecule-cost pairs generated by initially following policy $\pi$ until the game is ended from a multitude of target molecules. Chemical costs larger than 500 were extremely rare; those having estimates larger than 500 were truncated (see also Figure~\ref{fig:model_arch}). 

%%%%%%%%%%%%%%%%%%%%%%%%%%%%%%%%%%%%
%%%%%%%%%%%%%%%%%%%%%%%%%%%%%%%%%%%%

\subsection{Additional results}

\begin{figure} [ht]
\centering
\includegraphics[width=429pt]{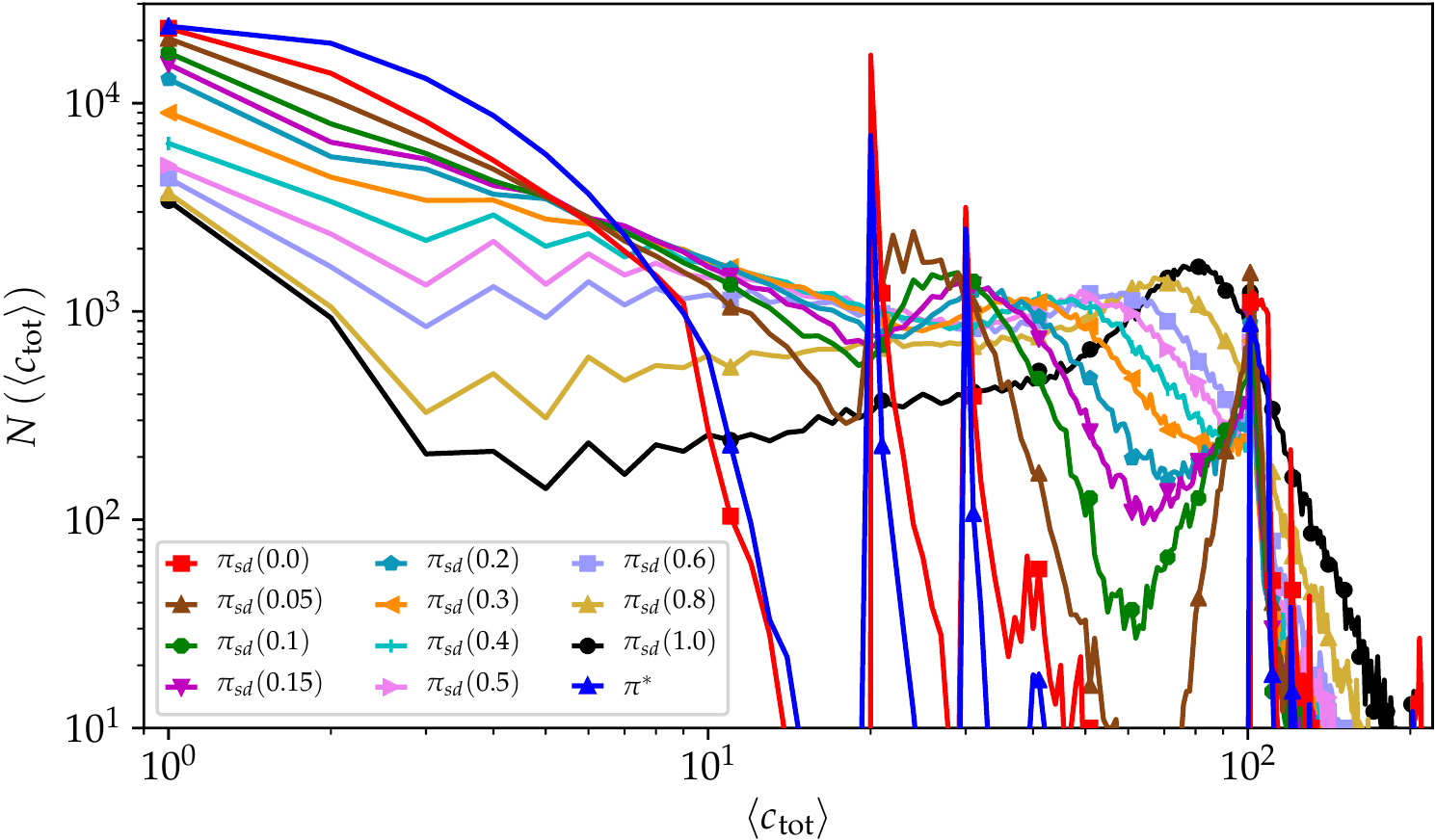}
\caption{The distribution of expected costs, $\langle c_\text{tot}\rangle$, over the set of 100,000 target molecules is shown for the greedy optimal policy $v^*(m)$ and the symmetric disconnection policy $\pi_{sd}$ for different values of $\varepsilon$.}
\label{fig:Sa}
\end{figure}

\begin{figure} [ht]
\centering
\includegraphics{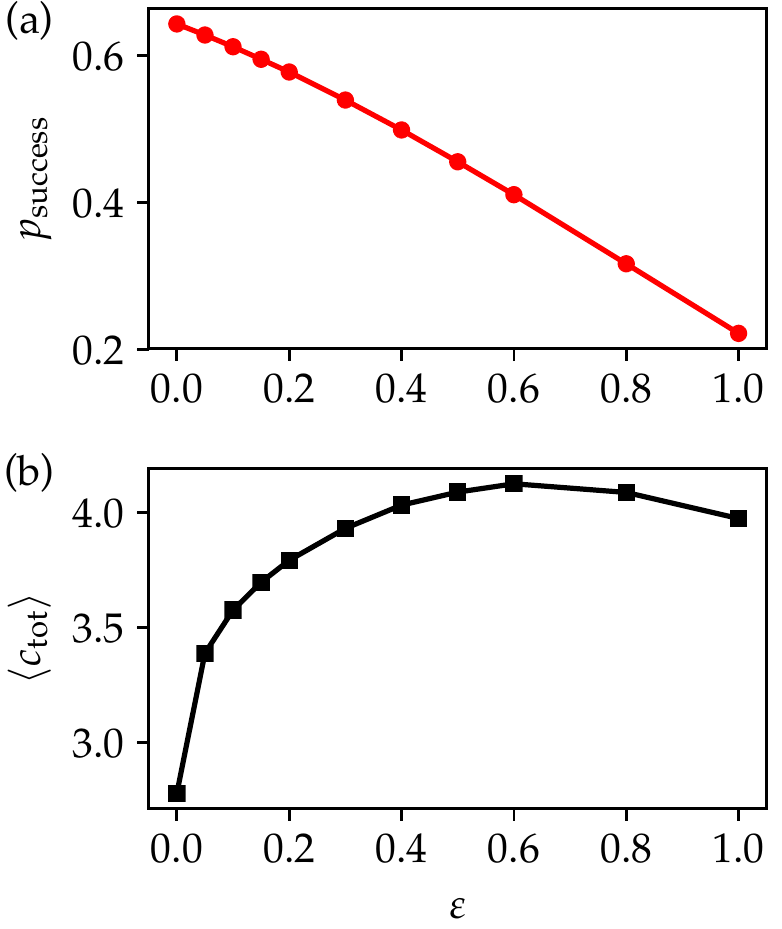}
\caption{(a) The probability of successfully synthesizing target molecules and (b) $\langle c_{\tot}\rangle$ (bottom) for those synthesized are shown versus $\epsilon$ for the symmetric disconnection policy $\pi_{sd}$.}
\label{fig:Sb}
\end{figure}

\begin{figure} [ht]
\centering
\includegraphics{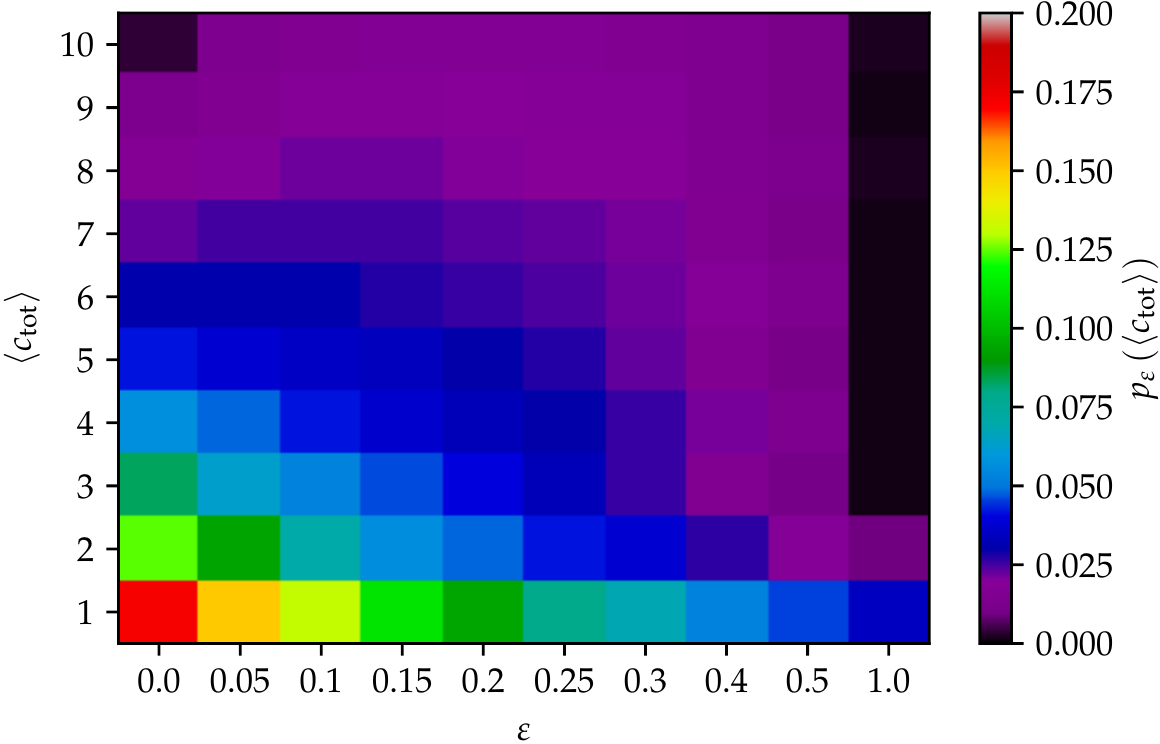}
\caption{Normalized probability distributions $p_\varepsilon(\langle c_{\tot}\rangle)$ are shown for different values of $\varepsilon$ for the symmetric disconnection policy $\pi_{sd}$. Note that the columns in the plot each sum up to one.}
\label{fig:Sc}
\end{figure}

\begin{figure} [ht]
\centering
\includegraphics{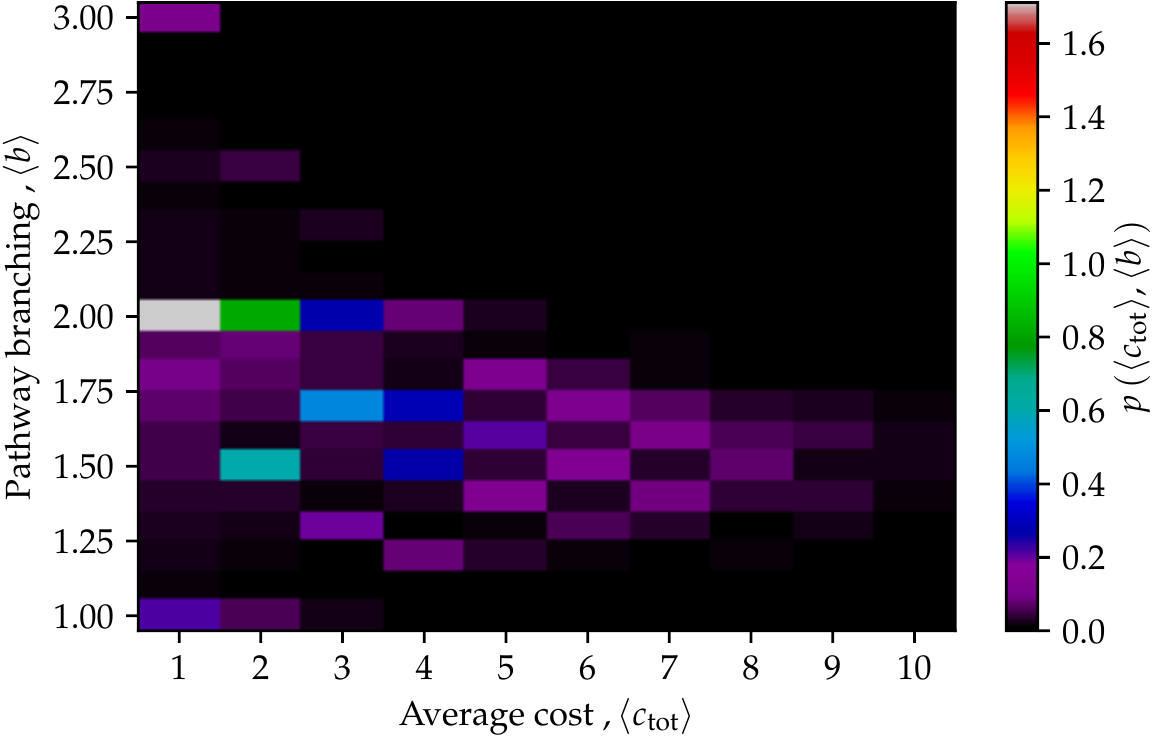}
\caption{A 2D histogram $p(\langle c_{\tot}\rangle, (\langle b_\text{tot} \rangle)$ is shown for the symmetric disconnection policy $\pi_{sd}$.}
\label{fig:Sd}
\end{figure}

Figure~\ref{fig:Sa} shows the full cost distribution for the heuristic policy ($\pi_{sd}$) and the learned optimal policy ($\pi^*$). Figure~\ref{fig:Sb}a shows the probability of successfully synthesizing a molecule from the training set ($p_\text{success}$) versus $\varepsilon$. Figure~\ref{fig:Sb}b shows the average cost ($\langle c_\text{tot} \rangle$) of the training set versus $\varepsilon$. Figure~\ref{fig:Sc} shows normalized probability distributions $p_\varepsilon(\langle c_\text{tot}\rangle)$ for the target costs in the training set for different values of $\varepsilon$. Figure~\ref{fig:Sd} shows the 2D probability distribution $p_\varepsilon(\langle c_\text{tot}\rangle)$ computed with $\varepsilon = 0$ for the training set.

%%%%%%%%%%%%%%%%%%%%%%%%%%%%%%%%%%%%
\begin{table}[t]
\renewcommand{\arraystretch}{1.3}
\centering
\begin{tabular}{ cccc|ccc } 

 \multicolumn{1}{c}{} & \multicolumn{3}{c|}{\textbf{Train} (100,00)} & \multicolumn{3}{c}{\textbf{Test} (25,000)} \\

 & $\pi_{sd}$ (\pct) & $\pi_*$ (\pct) & Tie (\pct) & $\pi_{sd}$ (\pct) & $\pi_*$ (\pct) & Tie (\pct)\\
 
 \cmidrule(lr){2-4}
 \cmidrule(lr){5-7}

 $c_\text{tot} < P_1$ & 0.4 & 35.0 & 47.9 & 6.1 & 21.3 & 47.9 \\
 $P_1 \le \text{tot} < P_2$ & 3.9 & 5.2 & 5.1 & 3.4 & 8.6 & 9.9 \\
 $c_\text{tot} \ge P_2$ & $<$1 & 1.6 & $<$1 & $<$1 & 1.1 & 1.3 \\
 Bulk & 4.4 & 41.8 & 53.8 & 9.9 & 31.0 & 59.1 \\ 
\bottomrule
 
\end{tabular}
\caption{\label{table_stats_supp} Comparison of the performance of $\pi_{sd}$ versus $\pi^*$. The values show the percent that one policy found a lower cost than the other, or whether the two policies found pathways with identical cost (e.g.\, a tie), for the molecules in the training and testing sets. All percents were computed using the size of the training ($\roughly$100,000) or testing set ($\roughly$25,000).} %Actual numbers - 95,774, 23,945. 
\end{table}
%%%%%%%%%%%%%%%%%%%%%%%%%%%%%%%%%%%%

%%%%%%%%%%%%%%%%%%%%%%%%%%%%%%%%%%%%
%%%%%%%%%%%%%%%%%%%%%%%%%%%%%%%%%%%%
\clearpage

\section{Acknowledgments}
This work was supported by the DARPA Make-It program under contract ARO W911NF-16-2-0023. We acknowledge computing resources from Columbia University's Shared Research Computing Facility project, which is supported by NIH Research Facility Improvement Grant 1G20RR030893-01, and associated funds from the New York State Empire State Development, Division of Science Technology and Innovation (NYSTAR) Contract C090171, both awarded April 15, 2010.

%%%%%%%%%%%%%%%%%%%%%%%%%%%%%%%%%%%%
%%%%%%%%%%%%%%%%%%%%%%%%%%%%%%%%%%%%

\bibliography{library.bib}

\clearpage

\end{document}